\pdfoutput=1

\documentclass{article}
\usepackage{spconf,amsmath,graphicx}
\usepackage{cite}
\usepackage{amsmath,amssymb,amsfonts}
\usepackage{algorithmic}
\usepackage{graphicx}
\usepackage{balance}
\usepackage{textcomp}
\usepackage{xcolor}
\usepackage{subfigure}
\usepackage{algorithm}
\usepackage{multirow}
\usepackage{url}
\usepackage{booktabs}
\usepackage{array}
\usepackage{enumitem,bm}
\def\method{CGNN}
\title{Learning on Graphs under Label Noise}
\name{Jingyang Yuan$^{1,2,*}$, Xiao Luo$^{3,*}$, Yifang Qin$^{1,2}$, Yusheng Zhao$^{1,2}$, Wei Ju$^{1,2,\dagger}$, Ming Zhang$^{1,2,\dagger}$}
\address{$^1$National Key Laboratory for Multimedia Information Processing, Peking University, Beijing, China\\
$^2$School of Computer Science, Peking University, Beijing, China \\
$^3$Department of Computer Science, University of California, Los Angeles, USA
\thanks{This paper is partially supported by grants from the National Key Research and Development Program of China with Grant No. 2018AAA0101902 and the National Natural Science Foundation of China (NSFC Grant Number 62276002). $^*$Equal contribution. $^{\dagger}$Corresponding authors (e-mail: juwei@pku.edu.cn; mzhang\_cs@pku.edu.cn). }
}

\begin{document}
%\ninept
%
\maketitle
\begin{abstract}
 Node classification on graphs is a significant task with a wide range of applications, including social analysis and anomaly detection. Even though graph neural networks (GNNs) have produced promising results on this task, current techniques often presume that label information of nodes is accurate, which may not be the case in real-world applications. To tackle this issue, we investigate the problem of learning on graphs with label noise and develop a novel approach dubbed Consistent Graph Neural Network (\method{}) to solve it. Specifically, we employ graph contrastive learning as a regularization term, which promotes two views of augmented nodes to have consistent representations. Since this regularization term cannot utilize label information, it can enhance the robustness of node representations to label noise. Moreover, to detect noisy labels on the graph, we present a sample selection technique based on the homophily assumption, which identifies noisy nodes by measuring the consistency between the labels with their neighbors. Finally, we purify these confident noisy labels to permit efficient semantic graph learning. Extensive experiments on three well-known benchmark datasets demonstrate the superiority of our \method{} over competing approaches.
\end{abstract}


\section{Introduction}

Graph neural networks (GNNs), which are based on the spectral graph theory, have shown promising results for a variety of graph machine learning problems, including node classification~\cite{tiwari2022exploring,li2019learning,kipf2017semi}, graph classification~\cite{ju2022ghnn,luo2022dualgraph,ju2022glcc}, and so on. Numerous GNNs have been presented in the literature to address the issue of node classification~\cite{xu2022heterogeneous}. These GNNs typically use message passing neural networks to learn efficient node representations. Despite their success, these approaches often presumptively believe that the labels of nodes on the graph are accurate, which would not be the case in practical applications like anomaly detection~\cite{zhong2019graph}. The performance of GNNs may be significantly hampered by these noisy nodes~\cite{song2022learning1}. To tackle this issue, we focus on the problem of node classification with label noise in this paper.

However, learning on graphs with label noise is a tough endeavor that requires us to address two key challenges:
(1) How to regularize GNNs to prevent noisy nodes from overfitting? GNNs are easy to overfit noisy nodes on the graph because of their higher model capacity. Therefore, a superior model generalization ability against label noise is expected by introducing a strong regularization factor.
(2) How to select noisy nodes in the graph? To ultimately eliminate label noise, noisy nodes should be chosen and their labels should be revised. Earlier approaches in computer vision often choose samples with large losses as noisy ones with the assumption that all the samples in the training set are independent and distributed uniformly~\cite{zhu2021second,cheng2020learning}. However, nodes are connected by a variety of edges on the graph, which makes it difficult to identify noisy nodes in our setting.

In this study, we offer a simple yet effective approach called Consistent Graph Neural Network (\method{}) for node classification with label noise. In particular, we take cues from graph contrastive learning~\cite{zhu2021graph,xu2021infogcl,hafidi2022negative,you2020graph} and propose a regularization term without label information, which produces two views through graph augmentation and then promotes two node views to have consistent representations compared to other nodes. When combined with supervised classification loss, this regularization term can help produce generalized node representations in the face of label noise.
In addition, we recall that the homophily assumption requires that linked nodes have consistent semantics. In this spirit, we offer a sample selection technique to find these noisy nodes on the graph, which identifies nodes with a large disagreement with their neighbors as a suspect. 
The semantic learning on the graph is then improved by relabeling a subset of confident noisy nodes. Extensive studies on three benchmark datasets show that our \method{} is superior to competing baselines.

All in all, this work contributes in the following ways: (i) We study an under-researched yet crucial problem of node classification with label noise and present a novel method named \method{} to address it. (ii) Our \method{} not only introduces a regularization term to provide consistent presentations under varied augmentations, but also detects nodes breaking the homophily assumption as noisy to clear the label noise. 
(iii) Extensive experiments on three widely used benchmark datasets show that our \method{} outperforms various state-of-the-art approaches.

\begin{figure*}[ht]
\centering
\setlength{\belowcaptionskip}{0cm}
\includegraphics[width=0.99\textwidth,keepaspectratio=true]{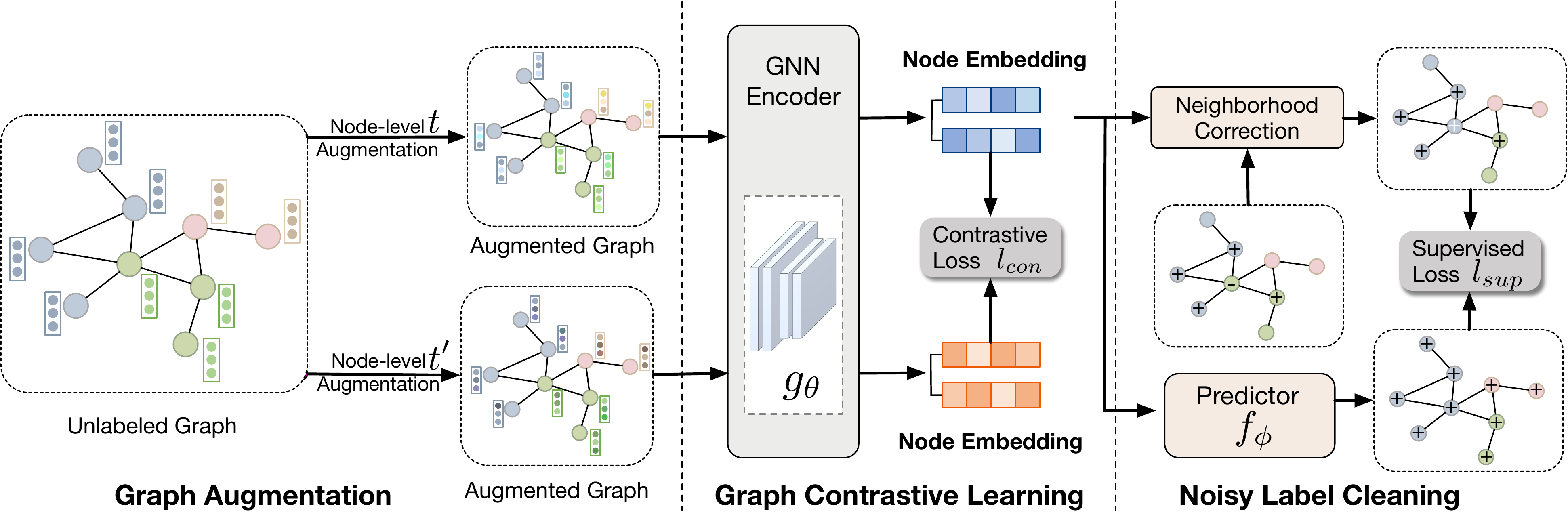}
\caption{Overview of our proposed \method{}.
}
\label{fig:framwork}
\end{figure*}

\section{Methodology}

% In this section, we first formalize our problem and then introduce our \method{} as shown in Fig. \ref{fig:framwork}. 

\subsection{Problem Definition}
We represent a graph as $\mathcal{G}=(\mathcal{V}, \mathcal{E})$ which consists of a set of $N$ nodes $\mathcal{V}$ and a set of edges $\mathcal{E} \subseteq \mathcal{V} \times \mathcal{V}$. Let $\bm{x}_i \in \mathbb{R}^d$ represent node attribute of node $i$ where $d$ is the dimension of attributes. Partial nodes $\mathcal{V}_L=\{1,\cdots,M\}$ have labels $\mathcal{Y}_L=\{{y}_1,\cdots, {y}_M\}$, which could be polluted by label noise. The aim of our problem is to predict the labels of $\mathcal{V}_U = \mathcal{V}/\mathcal{V}_L$ under label noise in $\mathcal{Y}_L$.

\subsection{Basic Architecture}
As shown in Fig. \ref{fig:framwork}, we choose the message passing neural network as the basic network architecture~\cite{xu2018powerful,kipf2017semi,velivckovic2017graph,dai2022towards}, where each node representation is updated by aggregating its neighborhood information. In formulation, let $\bm{h}_i^{(k)}$ denote the representation of node $i$ at layer $k$. The updating procedure can be formulated as follows:
\begin{equation}
\begin{aligned} \bm{p}_{i}^{(k)} &=\operatorname{AGGREGATE}^{(k)}\left(\left\{\bm{h}_{i}^{(k-1)}: j \in \mathcal{N}(i)\right\}\right) \\ \bm{h}_{i}^{(k)} &=\operatorname{COMBINE}^{(k)}\left(\bm{h}_{i}^{(k-1)}, \bm{p}_{i}^{(k)}\right) \end{aligned}
\end{equation}
in which $\mathcal{N}(i)$ represents the neighborhood set of node $i$. $\operatorname{AGGREGATE}^{(k)}(\cdot)$ and $\operatorname{COMBINE}^{(k)}(\cdot)$ denote the aggregation and combination operations at layer $k$, respectively. By stacking $K$ layers, we output the final representation $\bm{h}_i=\bm{h}_i^{(K)}$ for node $i$. 

\subsection{Graph Contrastive Learning as Regularization}

In most cases, the label information is used to guide the training of the message passing neural network.
However, the quality of node representations may drastically decline owing to label noise~\cite{dai2021nrgnn,jin2020graph,xu2020towards}. Here, we offer a new learning framework to address this issue, which uses graph contrastive learning to enforce consistent node representations under graph perturbation. By regularizing the learning process of node representations, our method improves the model generalization to resist the label noise. 

In detail, we first introduce two graph augmentation strategies where connect edges are randomly dropped and node attributes of partial nodes are randomly masked~\cite{zhu2021graph,jin2021multi,hafidi2022negative}. Then, two graph views can be obtained, i.e., $\mathcal{G}^1$ and $\mathcal{G}^2$. By applying the message passing neural network, we can output two views of representations for each node $i$, i.e., $\bm{h}^1_i$ and $\bm{h}^2_i$. Following the paradigm of contrastive learning, we enforce two node views to have consistent representations compared with other nodes. The contrastive learning objective of each pair $\left(\bm{h}^1_i, \bm{h}^2_i\right)$ is formulated as follows:

\begin{equation}
\ell\left(\bm{h}^1_i, \bm{h}^2_i\right)=-\log \frac{\exp \left(\operatorname{sim}\left(\bm{h}^1_i, \bm{h}^2_i\right)/\tau\right)}{\sum_{j=1}^{N} \exp \left(\operatorname{sim}\left(\bm{h}^2_i, \bm{h}^2_j\right)\tau\right)},
\end{equation}
where $\operatorname{sim}(\cdot, \cdot)$ denotes the cosine similarity between two vectors and $\tau$ is a temperature coefficient set to $0.5$ following \cite{you2020graph}. The final contrastive learning loss is formulated as:
\begin{equation}
    \mathcal{L}_{CL}= 
\frac{1}{2 N} \sum_{i=1}^{N}\left[\ell\left(\bm{h}^1_i, \bm{h}^2_i\right)+\ell\left(\bm{h}^2_i, \bm{h}^1_i\right)\right].
\end{equation}

In addition, we introduce a multi-layer perception (MLP) to output the prediction distribution for each node and the standard cross-entropy loss is employed to train labeled nodes on graphs. Let $\bm{q}_i = MLP(\bm{h}_i)$ denotes the output distribution for node $i$, and we have:
\begin{equation}\label{eq:sup}
   \mathcal{L}_{SUP} = -\frac{1}{|\mathcal{V}^{L}|} \sum_{i\in \mathcal{V}^{L}}\bm{y}_{i}^{T}\log\bm{q}_{i}.
\end{equation}
where $\bm{y}_{i}$ denotes the one-hot label embedding of node $i$. 

Finally, we combine both the supervised learning loss with the contrastive regularization term. The total loss is formulated as follows:

\begin{equation}\label{eq:total}
    \mathcal{L} = \mathcal{L}_{CL}+ \mathcal{L}_{SUP}.
\end{equation}

\subsection{Neighborhood-based Label Noise Correlation} 

In this part, we want to locate noisy nodes on the graph to update their labels, and then apply the new labels to the graphs for more efficient node classification. In our case, the identification of noisy nodes is complicated by the fact that the nodes in the network are linked by a wide variety of edges~\cite{song2022learning}. As a means of dealing with this, we draw on the homophily assumption, which states that the semantics of adjacent nodes should be similar. Thus, we identify those nodes whose predictions greatly differ from those of their neighbors as noisy ones and then revise the confident signals by comparing the similarity of node representations.

In detail, we first obtain the pseudo-label of every unlabeled node, i.e., $\hat{y}_i$. Then the consistency score is calculated by comparing the label of each node in $\mathcal{V}_L$ with their neighborhood. Here, the most common label in the neighborhood is derived by the following formulation:
\begin{equation}
    c_i=max_{c}\{ \# \{j |y^*_j=c,j\in \mathcal{N}(i) \} \},
\end{equation}
If $c_i\neq y_i$, we have to revise the node label based on neighborhood information.
However, neighboring nodes could still be dissimilar to the central node as an exception in large-scale datasets. To tackle this, we introduce a similarity consistency score to calculate the proportion of nodes with similar representations in the embedding space. The score is formulated as follows:
\begin{equation}
    a_i =  \frac{\# \{j |y^*_j=c_i, sim(\bm{h}_j,\bm{h}_i)>\gamma ,j\in \mathcal{N}(i) \}}{\# \{j |y^*_j=c_i,j\in \mathcal{N}(i) \}},
\end{equation}
where $y^*_j = y_j$ if $j$ is labeled or $\hat{y}_j$ otherwise and $\gamma$ is a predefined threshold.

Then a threshold $\omega$ is utilized to cut off the similarity consistency score. The confident noisy nodes with both large similarity consistency scores will be updated as follows:
\begin{equation}
    \tilde{y}_i=\left\{\begin{array}{ll}
  c_i & y_i \neq c_i \wedge a_i>\omega , \\
  y_i & otherwise ,   \end{array}\right.
\label{equ:similarity_1}
\end{equation}
From Eq. \ref{equ:similarity_1}, the labels are kept if their neighbourhood suggested label is the same or the similarity consistency score is above the threshold. Then we revise the Eq. \ref{eq:sup} into: 
\begin{equation}\label{eq:sup_update}
   \mathcal{L}_{SUP} = -\frac{1}{|\mathcal{V}_{L}|} \sum_{i\in \mathcal{V}_{L}}\tilde{\bm{y}}_{i}^{T}\log\bm{q}_{i}.
\end{equation}
where $\tilde{\bm{y}}_{i}$ denotes the one-hot label vector from $\tilde{y}_i$. 
In this way, we identify and clean confident noisy nodes, thus facilitating effective semantic learning on graphs.  

During optimization, we first warm up the message passing neural network using Eq. \ref{eq:total}. Then we iteratively remove label noise every $R$ epochs and train the graph neural network using revised labels. The whole network is optimized using the stochastic gradient descent algorithm.

\begin{table}[t]
\renewcommand\arraystretch{0.935}
    \centering
    \setlength{\abovecaptionskip}{0.2cm}    
    \caption{Details of the datasets.}
    \label{tab:datasets}
    \resizebox{0.48\textwidth}{!}{
    \begin{tabular}{ccccc}
        \toprule
        \midrule 
        Datasets & Nodes& Edges& Features&  Classes \\
        \midrule
        Amazon Photo & 7,650 & 119,081 & 745 & 8 \\
        Amazon Computers & 13,752 & 245,861  & 767 & 10 \\
        Coauthor CS & 18,333 & 81,894 & 6,805 & 15 \\

        \bottomrule
    \end{tabular}
    }
\vspace{-0.5cm}
\end{table}

\section{Experiments}
In this section, we present the experimental results to prove the effectiveness of \method{}. We first introduce the datasets and baselines and corresponding settings, followed by experimental results and related analysis.
\subsection{Experiment Setup}
\textbf{Datasets and Settings.} We conduct experiments on three famous network datasets, which are two co-purchase networks from Amazon (Amazon Computers and Amazon Photo) as well as one co-author network dataset (Coauthor CS). For all three datasets, we randomly choose a fixed part of node labels for supervising. Since the origin training labels in the datasets are clean, we use two corruption manners to make label noise. The first manner is \textbf{Uniform Noise}, which uniformly flips a label to other classes with a probability of $p$. The second manner is \textbf{Pair Noise}, which flips a label just to its most similar class in the probability of $p$. The pair noise manner assumes that labels tend to be misled to their most similar pair classes. The details of three datasets are shown in Table \ref{tab:datasets}.

\noindent\textbf{Baselines and Protocols.} We compare our method with the state-of-art GNNs and methods on noisy-resistant learning, including GCN~\cite{kipf2017semi}, GIN~\cite{xu2018powerful}, Forward~\cite{patrini2017making}, Coteaching+ \cite{yu2019does}, D-GNN~\cite{nt2019learning}, CP~\cite{zhang2020adversarial}. We provide the mean accuracy and standard deviations of 5 runs for each baseline. In our experiments, the training label rate is set to 0.01 and the percent of noisy labels is fixed at $20\%$.

\noindent\textbf{Implementation Details.} 
All the methods are implemented using PyTorch Geometric. 
Our \method{} uses a message passing neural network with three layers as our backbone, and the hidden dimension is set to 256. The noise removal hyperparameters $\gamma$ and $\omega$ are both tuning from candidates $\{0.6, 0.7, 0.8, 0.9, 0.95\}$ with details in Sec. \ref{sec:sc} and we set them to $0.8$ as default. 

\begin{table*}[htbp]
\renewcommand\arraystretch{0.935}
    \centering
    \setlength{\abovecaptionskip}{0.2cm}    
    \caption{Main results (Accuracy(\%)±Std) on three datasets under various types of label noise. }
    \label{tab:main_results}
    \resizebox{0.95\textwidth}{!}{
    \begin{tabular}{c|c|cccccccc}
        \toprule
        \midrule 
        Dataset& Noise& GCN& GIN&  Forward& Coteaching+& D-GNN& CP& Ours& \\
        \midrule
        \multirow{2}*{Co-author CS} & Uniform &
        80.3 ±1.4 &80.5 ±1.2 &81.3 ±0.4 &80.7 ±1.4 &81.9 ±1.1 &82.5 ±1.2 & \textbf{84.1} ±\textbf{0.4} \\
        & Pair &     
        79.5 ±1.1 &79.3 ±1.7 &80.5 ±0.8 &77.6 ±3.3 &79.9 ±0.7 &80.2 ±1.0 & \textbf{81.0} ±\textbf{1.1}\\
        \midrule 
        \multirow{2}*{Amazon Photo} & Uniform &
        82.2 ±0.9 &82.1 ±0.4 &79.8 ±0.6 &78.5 ±0.6 &84.5 ±3.4 &83.7 ±0.7 & \textbf{85.3} ±\textbf{0.9}\\
        & Pair &
        80.9 ±1.2 &81.2 ±1.5 &78.7 ±0.3 &75.5 ±1.8 &83.2 ±2.2 &83.1 ±1.1 & \textbf{85.1} ±\textbf{0.7}\\
        \midrule 
        \multirow{2}*{Amazon Computers} & Uniform &
        76.9 ±1.2 &76.4 ±0.7 &78.4 ±0.4 &70.3 ±3.6 &81.2 ±0.6 &80.6 ±1.3 & \textbf{81.8} ±\textbf{0.3}\\
        & Pair &
        76.3 ±1.6 &76.8 ±1.2 &77.3 ±0.6 &60.9 ±3.9 &81.3 ±1.0 &80.4 ±2.2 & \textbf{81.6} ±\textbf{0.2}\\

        \bottomrule
    \end{tabular}
    }
    \vspace{-0.3cm}
\end{table*}

\subsection{Performance Evaluation}

The comparison of our \method{} and baselines are shown in Table \ref{tab:main_results}. We can observe that (i) GCN and GIN achieve the worst performance in the listed baselines. That's because these two supervised models are affected seriously by noised labels. That shows the necessity of studying node classification with label noise. (ii) Two robust learning models Coteaching+ and D-GNN still perform much worse performance compared with our \method{}, which shows that designing effective methods for learning on graphs with label noise is challenging. (iii) Our proposed \method{} performs best on all datasets and noises. To be specific, \method{} outperforms the best baseline by 1.4\%, and 1.9\% on Co-author CS and Amazon Photo.
Also, \method{} has the least standard deviation compared with other methods. That means our \method{} has a more stable performance in different training situations.

\begin{table}[htbp]
\renewcommand\arraystretch{0.935}
    \centering
    \setlength{\abovecaptionskip}{0.2cm}    
    \caption{Ablation study on Amazon Photo dataset.}
    \label{tab:ablation}
    \resizebox{0.48\textwidth}{!}{
    \begin{tabular}{c|ccccc}
        \toprule
        \midrule 
         Noise& \method\;w/o contr& \method\;w/o corr&  Full Model& \\
        \midrule
        Uniform & 84.4 ±0.6 &85.1 ±0.9 & \textbf{85.3} ±\textbf{0.9}\\
        Pair & 84.3 ±0.7 &85.0 ±0.9 & \textbf{85.1} ±\textbf{0.7}\\

        \bottomrule
    \end{tabular}
    }
\vspace{-0.5cm}
\end{table}

\subsection{Ablation Study}
We conduct ablation studies to find out the importance of each part in our proposed \method{}. To investigate whether contrastive learning is useful to extract node features, we remove the contrastive learning objective in Eq. \ref{eq:total}. We name the variant as \method{} w/o contr. Also, in order to demonstrate the effectiveness of neighborhood-based correlation, we train another variant \method{} w/o corr by directly sending noisy labels for supervision. The results are reported on uniform noise and pair noise corrupted Amazon Photo dataset. Noisy rate is set to 20\%. For each ablation experiment, we train five runs and record their average performance and the standard deviation. The comparison of \method{} w/o contr, \method{} w/o corr and \method{} are shown in Table \ref{tab:ablation}. We can see a performance decreasing with \method{} w/o remo, which means label correction helps the model to avoid noise misleading. We also detect similar performance degradation in \method{} w/o contr, indicating contrastive learning is useful in representation learning.

\begin{figure}[htbp]
\centering

\subfigure[Uniform noise]{
\begin{minipage}[t]{0.225\textwidth}
\centering
\includegraphics[width=3.5cm]{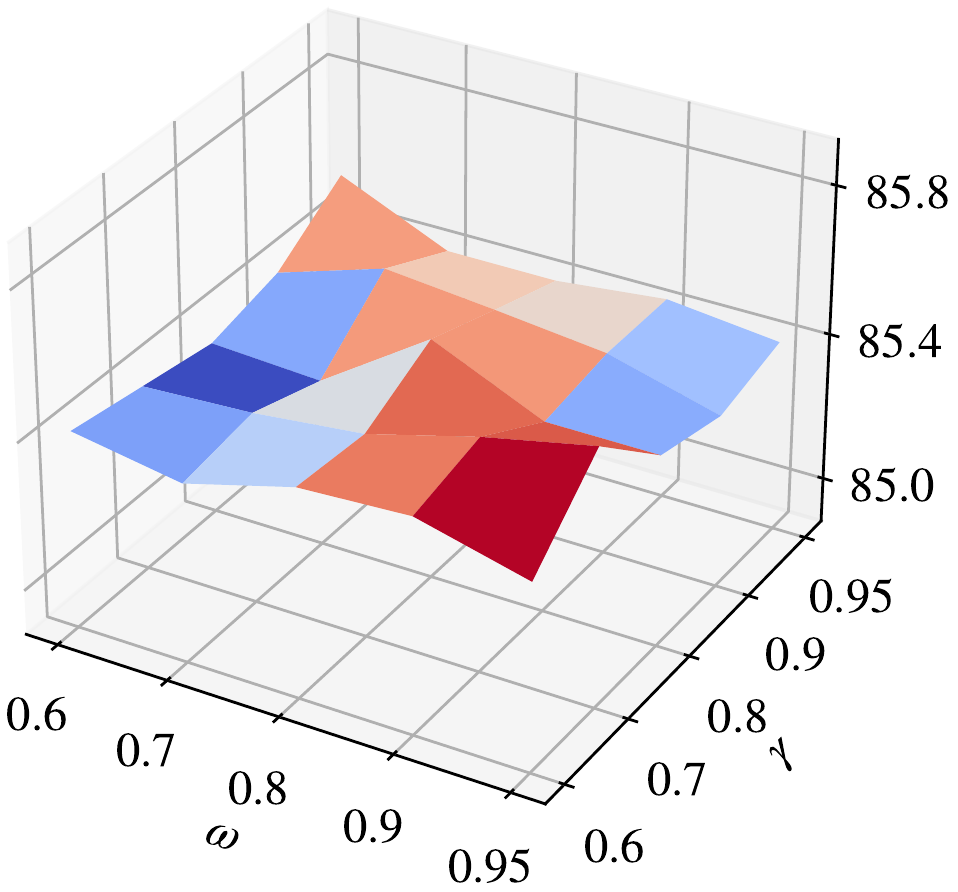}
\end{minipage}
}
\subfigure[Pair noise]{
\begin{minipage}[t]{0.225\textwidth}
\centering
\includegraphics[width=3.5cm]{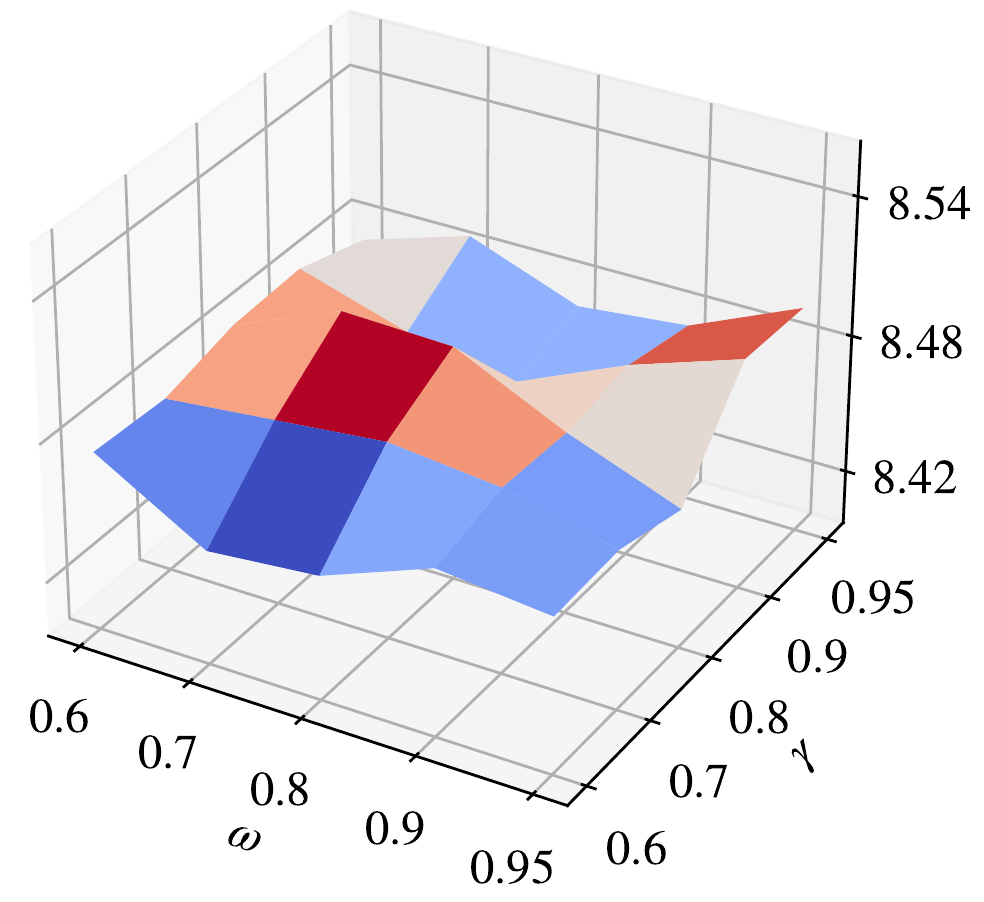}
\end{minipage}
% \caption{dd}
}
\vspace{-0.3cm}
\caption{Parameter sensitivity analysis on Amazon Photo.}
\vspace{-0.4cm}
\label{fig:sensi}
\end{figure}

\subsection{Hyperparameter Sensitivity Analysis}\label{sec:sc}
Here we study how the results of our \method{} are influenced by two hyperparameters $\omega$ and $\gamma$, which control neighborhood-based noise removal. $\gamma$ is a threshold to measure whether the cosine similarity of two linked nodes is high enough. $\omega$ is the neighborhood voting threshold, and only if the ratio of a similar neighbor with the same label to all neighbors with the same label is higher than $\omega$, the node label will be corrected. We select $\omega$ and $ \gamma$ from $\{0.6, 0.7, 0.8, 0.9, 0.95\}$. We report the results on Amazon Photo dataset with uniform and pair noise, the noise rate is also set as 20\%.
The experiment results are demonstrated in Fig.\ref{fig:sensi}. We note that the maximum variation in the graph is smaller than 1\%, indicating the model is quite stable. Though hyperparameters affect model performance slightly on uniform noise corrupted data, we can observe on pair noise corrupted data that the too large $\gamma$ or too low $\omega$ higher decreases the performance. So we set $\gamma, \omega$ both to 0.8 for best prediction accuracy.

\section{Conclusion}
\vspace{+0.15cm}
This paper studies a underexplored problem named node classification with label noise and propose a novel method named \method{} to solve it. Our \method{} uses graph contrastive learning as a regularization term to encourage consistency in node representations between two graph views. In addition, we provide a noisy sample selection strategy based on the homophily assumption to explore the inconsistency between node labels and with the labels of its neighbours. Extensive experiments on three widely used benchmark datasets show that our \method{} is better than competing state-of-the-arts.


\bibliographystyle{IEEEbib}
\bibliography{refs}

\end{document}